\documentclass[11pt]{article}

\usepackage[final]{acl}

\usepackage{times}
\usepackage{latexsym}
\usepackage{tabularx}

\usepackage[T1]{fontenc}
\usepackage[utf8]{inputenc}

\usepackage{microtype}

\usepackage{inconsolata}

\usepackage{graphicx}
\usepackage{multirow}
\usepackage{booktabs}
\usepackage{tabularx}
\usepackage{caption}
\usepackage{subcaption}
\usepackage{balance}

\title{Better as Generators Than Classifiers: Leveraging LLMs and Synthetic Data for Low-Resource Multilingual Classification}

\author{Branislav Pecher$^\dagger$, Jan Cegin$^\dagger$, Robert Belanec$^{\spadesuit}$$^\dagger$, \\ {\bf Ivan Srba$^\dagger$, Jakub Simko$^\dagger$, Maria Bielikova$^\dagger$} \\
$^\dagger$ Kempelen Institute of Intelligent Technologies, Bratislava, Slovakia\\
$^{\spadesuit}$ Faculty of Information Technology, Brno University of Technology, Brno, Czechia \\
\texttt{\{name.surname\}}@kinit.sk\\ }

\begin{document}
\maketitle
\begin{abstract}
Large Language Models (LLMs) have demonstrated remarkable multilingual capabilities, making them promising tools in both high- and low-resource languages. One particularly valuable use case is generating synthetic samples that can be used to train smaller models in low-resource scenarios where human-labelled data is scarce. In this work, we investigate whether these synthetic data generation capabilities can serve as a form of distillation, producing smaller models that perform on par with or even better than massive LLMs across languages and tasks. To this end, we use a state-of-the-art multilingual LLM to generate synthetic datasets covering 11 languages and 4 classification tasks. These datasets are then used to train smaller models via fine-tuning or instruction tuning, or as synthetic in-context examples for compact LLMs. Our experiments show that even small amounts of synthetic data enable smaller models to outperform the large generator itself, particularly in low-resource languages. Overall, the results suggest that LLMs are best utilised as generators (teachers) rather than classifiers, producing data that empowers smaller and more efficient multilingual models.
\end{abstract}

\section{Introduction}

\begin{figure}[tbh]
    \centering
    \includegraphics[width=0.99\linewidth]{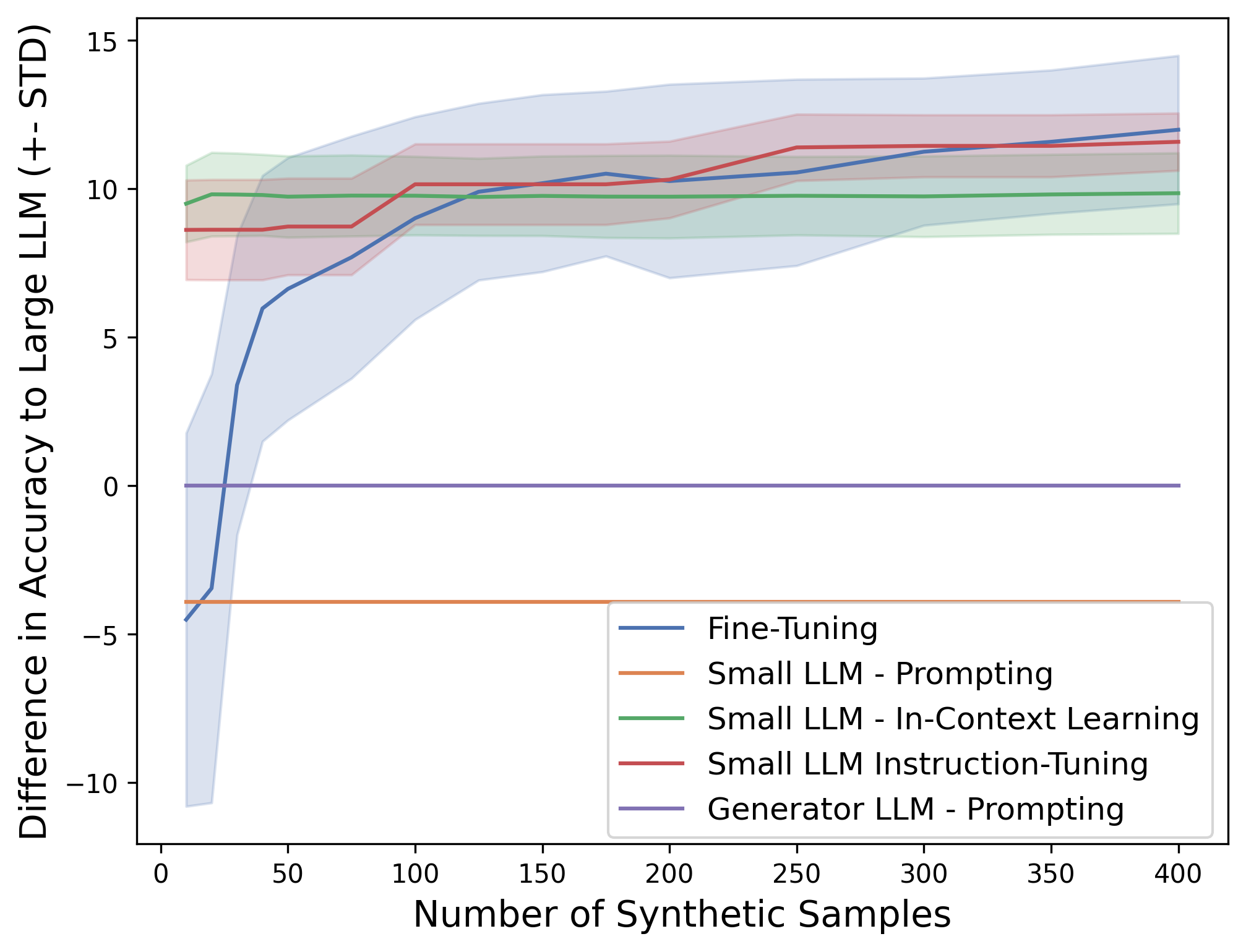}
    \caption{Smaller models trained with synthetic samples often outperform the large language model used for generating such samples. The results obtained by 5 models are aggregated over 11 languages and 4 tasks. The small LLM results are aggregated over 3 models.}
    \label{fig:introduction}
\end{figure}

% With the rise of large language models (LLMs) based on the transformer architecture, there has been a significant increase in the multilingual capabilities of language models \citep{}. This also caused the language models to scale up to relatively high amounts of trainable parameters and high amounts of required training data for pre-training and fine-tuning. This requirement for large amounts of training data discriminates against low-resource languages and makes the LLMs underperform in such languages. In addition, the performance of many state-of-the-art LLMs on tasks in non-English high-resource languages is often subpar to tasks in the English language. This underperformance often correlates with the size of the model. Therefore, there is a potential for synthetic sample generation to increase the size of training data.

Large Language Models (LLMs) have emerged as powerful generators of synthetic data, enabling the training of smaller, efficient downstream models across a wide range of tasks. It was successfully applied to different tasks, often rivalling or even surpassing traditional augmentation methods~\citep{piedboeuf-langlais-2023-chatgpt, ONAN2023101611, cegin-etal-2023-chatgpt, sen-etal-2023-people}. Importantly, recent work has shown that LLMs are capable of producing high-quality synthetic data for low-resource languages~\citep{anikina2025rigorousevaluationllmdata}. %For example, LLM-based generation has been explored for multilingual question answering~\citep{kramchaninova-defauw-2022-synthetic, namboori2023gemquad}, fact-checking in under-represented languages~\cite{chung2025beyond}, and classification tasks~\cite{anikina2025rigorousevaluationllmdata}. As such, they are not only capable of generating high-quality data for high-resource languages, but for a variety of low-resource languages when used for downstream model fine-tuning.

Besides generation, the LLMs are frequently used directly as classifiers through zero or few-shot prompting~\citep{brown2020language, sun2023pushing, dong2022survey, pecher2023effects}. A large body of work has examined their performance in this role, comparing them against established supervised classification models (obtained through fine-tuning) from different perspectives. When faced with a severe lack of labelled samples (e.g., having only around $4-64$ samples), the zero and few-shot prompting often outperforms fine-tuning of smaller models~\citep{ma-etal-2023-large, hongjin2022selective, logan-iv-etal-2022-cutting, gao-etal-2021-making}. However, with enough labelled samples (depending on the task, it can be as low as $100$~\citep{pecher2024comparing}), the smaller models obtained through fine-tuning or instruction tuning often achieve performance on par or better than the large language models~\citep{schick-schutze-2021-just, qin2023chatgpt}. However, the behaviour in low-resource languages and other settings is not well understood, as the majority of studies focus on English and other high-resource languages.
%Similarly, when the comparisons are done in a fair manner (comparing models of similar sizes), the zero and few-shot prompting often achieves worse performance~\citep{mosbach-etal-2023-shot}. Moreover, the different studies usually do not consider the computational efficiency or the sensitivity to the effects of randomness~\citep{pecher2024comparing, mosbach_stability_2021, dodge2020fine, le-scao-rush-2021-many}. Finally, the behaviour in low-resource languages and other settings is not well understood, as the majority of the studies focus on English and other high-resource languages.

In this work, we shift the perspective and instead of focusing on large language models as end-task classifiers, we ask whether we \textbf{can leverage their strong generation capabilities as a form of data-driven distillation in low-resource settings.} Specifically, we investigate whether the \textit{synthetic samples} generated by the large multilingual language models can be used to improve smaller models to achieve comparable or even superior performance in low-resource multilingual classification. Namely, we investigate three options for training or improving small models: 1) through \textit{fine-tuning} with synthetic samples; 2) \textit{in-context learning with smaller LLMs} using synthetic samples as in-context examples; and 3) \textit{instruction-tuning of smaller LLMs} with synthetic samples. Afterwards, we systematically compare such small models with the large model used for generating the synthetic samples across \textit{11 typologically diverse languages} (from low-, medium- and high-resource language groups) and \textit{4 classification tasks}. To provide a more in-depth analysis, we also investigate how the comparison changes as the number of available synthetic samples increases and how the sensitivity to the effects of randomness affects it. The aggregated results are presented in Figure~\ref{fig:introduction}. Overall, within our experiments, we perform over 30,000 fine-tuning and instruction-tuning of the smaller models. Our main contributions and findings are\footnote{To support replicability, we provide the source code of our experiments at \url{https://github.com/kinit-sk/multilingual-classifiers-not-generators}}:
\begin{itemize}
    \item We present a comprehensive analysis of the role of LLMs as generators of synthetic samples for low-resource classification, which can be used to train more efficient (in terms of performance and computational costs) smaller models. We perform the analysis across 11 languages and 4 classification tasks, comparing the large generation model with fine-tuning of a multilingual encoder and 3 smaller LLMs, which are used through prompting, in-context learning, and instruction-tuning.
    \item We show that even small amounts ($20$ on average for low- and medium-resource, and $100$ for high-resource languages) of synthetic samples allow smaller models to match or outperform the large generator model itself, especially in low-resource settings. At the same time, performance in high-resource settings is more task-dependent, with popular and overly represented tasks often showing no performance benefit from synthetic samples, only a reduction in computation by having a smaller model. As such, it is more efficient to utilise the LLMs as generators of synthetic samples in low-resource languages that can be used to empower smaller models rather than using them as classifiers.
    \item The synthetic samples can help achieve similar performance to human-labelled samples in severely resource-constrained settings, but when increasing the number of available samples, their benefit quickly stagnates in comparison to the human-labelled samples, due to lower diversity and informativeness of the generated samples.
\end{itemize}

\section{Related Work}

Since their emergence, Large Language Models (LLMs), such as GPT-4 and Llama, have become central tools for generating and augmenting synthetic data. This strategy has proven effective across diverse applications, including automated scoring~\cite{fang2023using}, intent classification~\cite{sahu-etal-2022-data}, sentiment analysis~\cite{piedboeuf-langlais-2023-chatgpt, ONAN2023101611, yoo-etal-2021-gpt3mix-leveraging}, hate speech detection~\cite{sen-etal-2023-people}, news categorization~\cite{piedboeuf-langlais-2023-chatgpt}, and personalised recommendation~\cite{contect-based-recom}

Although much of this progress has been centred on English, recent studies highlight the growing use of LLMs for synthetic data creation in low-resource languages. Multilingual generation has been employed to build datasets for question answering~\cite{kramchaninova-defauw-2022-synthetic, namboori2023gemquad, putri-etal-2024-llm}, fact-checking~\cite{chung2025beyond}, commonsense reasoning in culturally nuanced settings~\cite{pranida2025syntheticdatagenerationculturally}, and cross-lingual named entity recognition~\cite{liu-etal-2021-mulda}. Similarly, LLM-based data augmentation has supported sentiment stance detection~\cite{ZOTOVA2021114547} and classification tasks in diverse languages~\cite{glenn-etal-2023-jetsons, anikina2025rigorousevaluationllmdata}.

Multiple studies have already focused on comparing smaller models that are further improved via fine-tuning or instruction-tuning, with larger general models, which are used through prompting or in-context learning, on text classification tasks~\cite{dong2022survey, liu2023pretrain-prompt-survey}. Many studies find that with enough labelled samples, the smaller models can outperform the significantly larger ones~\cite{schick-schutze-2021-just, qin2023chatgpt, lehman2023we, pecher2024comparing}. The number of required samples to achieve this superior performance is often quite small, in the range between $100 - 1000$ samples~\citep{pecher2024comparing}. Only in the extremely resource-constrained setting, where the models are fine-tuned with the same number of samples as are used for in-context learning (e.g., $4-64$), the zero and few-shot prompting is able to outperform the smaller models~\citep{ma-etal-2023-large, hongjin2022selective}. However, when focusing on a fairer comparison, where the models of similar sizes are used, the fine-tuning (or instruction-tuning) often leads to significantly better performance~\citep{mosbach-etal-2023-shot}. All of the previous works perform the comparison on English or other high-resource languages. When moving to more low-resource settings, either using low-resource languages or tasks that lack labelled samples, the situation is often different. The small models can often outperform much larger LLMs even under the same data budgets, as the in-context learning often falters and is characterised by significant performance variance~\citep{asai-etal-2024-buffet, micallef2025melabenchv1}. Similarly, when human-labelled samples are lacking, generating synthetic samples for question answering can lead to high performance, sometimes even outperforming the original training sets~\citep{puri-etal-2020-training}.

However, to the best of our knowledge, there are no works that investigate whether the large language models are better as generators (a task they were designed for) than classifiers, especially in a low-resource setting. In this paper, we fill this gap and ask whether we can use the large language models to generate synthetic samples in the low-resource languages that are then used to empower (e.g., using them for fine-tuning or as in-context examples) smaller models and lead to more efficient classification (in terms of performance and computation costs).

\section{Methodology: Data-driven Distillation Through Synthetic Samples}

We analyse whether it is more efficient (for classification in low-resource settings) to use the large language models as generators of synthetic samples instead of classifiers. In essence, we aim to distil the LLM into smaller models that require lower computational resources, while maintaining or even surpassing the performance of the original LLM. To achieve this, we perform two steps: 1) generating synthetic samples; and 2) using the samples for training or as in-context examples.

\paragraph{Generating Synthetic Samples}

To generate the synthetic samples, we use the current state-of-the-art approaches for various languages identified in~\citet{anikina2025rigorousevaluationllmdata, cegin2024userandomselectionnow}. The studies found that the generation of synthetic samples works significantly better when presenting the model with a few representative samples for each label and language. Therefore, we randomly sample 10 examples from the train data and include them in the prompt given to the LLM as examples from the dataset to guide the generation. Collecting such a number of human-labelled samples is also achievable in low-resource settings, while not producing as large a performance improvement as the synthetic samples that are generated using these samples. The prompt used for this generation is included in Appendix~\ref{app:further-details}. As part of the generated samples may be problematic, such as those of lower quality or containing different languages (mainly English), we apply self-revision as a filtering mechanism. In this step, the generator LLM is used as a judge, by presenting it with the task description, the generated synthetic sample, and the criteria (e.g., containing only text in the specified language), and asking whether the sample is acceptable. Besides relying solely on the LLM for this task, we also manually check a subset of the samples after generation, and if necessary, generate more. We generate samples until we have a total of 200 samples for each label present in the dataset.

\paragraph{Using Synthetic Samples in Smaller Models} After obtaining the synthetic samples, we use them to train smaller models on the given tasks and languages. First, we use them to train a multilingual classification model through \textit{fine-tuning}, where we initialise a pretrained model, add a classification head with dropout, and train all of the layers with a small learning rate. Second, we randomly select a set of these synthetic samples to be used as in-context examples for smaller LLMs. Third, we use synthetic samples to instruction-tune the smaller LLM as well (using LoRA) and then use it with in-context learning. The fine-tuning approach represents the cheapest option in terms of computational resources, while instruction-tuning is the most expensive. Finally, we also compare all of these approaches with a simple zero-shot prompting with small LLMs and the large LLM used for generating the samples. 

As we are also interested in analysing how the behaviour of the models changes as we use more synthetic samples, we repeat the whole process of training or improving the small models across different sizes of the generated samples. We randomly subsample 200 generated samples into subsets of different sizes, starting with 10 samples (ensuring we have at least 2 samples per class), up to the full set of samples. Furthermore, we are interested in the impact of this random subsampling, and other sources of randomness, such as a random initialisation of the models or a choice of in-context examples that were observed to cause significant variance in results in previous works~\cite{sclar_quantifying_2023, weber-etal-2023-mind, pecher2023effects, pecher2024comparing, gundersen2022sources_of_irreproducibility}. As such, we run all experiments 20 times, reporting the mean accuracy and standard deviation across these runs (unless specified otherwise). More detailed setup for the models, such as prompt format, learning rate, and the sizes of the dataset subsets, is provided in Appendix~\ref{app:further-details}.

\section{Experiments}

\paragraph{Languages and Datasets} We use 11 typologically diverse languages belonging to different groups based on their sizes. From the low-resource languages, we use Azerbaijani, Romanian, Slovenian, Telugu, and Welsh. For the medium-resource languages, we use Hebrew, Indonesian, Swahili, and Thai. Finally, for the high-resource languages, we use English and German. We evaluate all of these languages in three classification tasks of different characteristics, such as their difficulty: 1) sentiment classification; 2) topic classification; and 3) intent recognition. In addition, we evaluate English on sarcasm detection to further explore the benefit of synthetic samples not only on low-resource languages, but also on low-resource tasks in high-resource languages.

For \textit{sentiment classification}, containing two labels, no single multilingual dataset currently exists. We therefore use the \textbf{combined ten datasets} for low-resource languages from~\cite{gurgurov-etal-2025-gremlin, gurgurov-etal-2024-adapting} with two additional datasets for English and German from~\cite{mollanorozy-etal-2023-cross}. This combination was first introduced in~\cite{anikina2025rigorousevaluationllmdata}. These datasets vary in coverage and text domain: the German set focuses on transportation and infrastructure, while the Romanian data primarily consists of product reviews. For \textit{topic classification}, we rely on \textbf{SIB-200}~\cite{adelani-etal-2024-sib}, which provides seven topic labels derived from the FLORES-200 machine translation corpus, annotated at the sentence level. For \textit{intent recognition}, we utilise the \textbf{MASSIVE} dataset \cite{fitzgerald-etal-2023-massive}, a multilingual benchmark for virtual assistant evaluation. We restrict the label set to ten distinct, most common intents to avoid possible semantic overlaps. Finally, for \textit{sarcasm detection}, treated as binary classification in this paper, we use the \textbf{SemEval 2021 Task 7}~\cite{meaney-etal-2021-semeval} data used for humour, offence, and sarcasm detection.

\paragraph{Models} For generating the synthetic samples, we use the \textbf{LLaMA-3 70B}~\citep{grattafiori2024llama} instruction-tuned large language model. We chose this model as it is open-source (and does not face problems with reproducibility, unlike many of the closed-source models) and has shown the best generation capabilities in our preliminary experiments. When comparing with other models, we use the model in a zero-shot setting. For the smaller models, we use \textbf{XLM-RoBERTa Large}~\citep{conneau-etal-2020-unsupervised} for fine-tuning (due to its multilingual capabilities) and \textbf{LLaMA-3.1-8B}~\citep{grattafiori2024llama}, \textbf{Gemma-3-4B}~\citep{gemma_2025}, and \textbf{Qwen-2.5-7B}~\citep{qwen2.5} instruction-tuned large language models for prompting, in-context learning, and instruction-tuning. 

\subsection{Comparison Across Language Groups}
\label{sec:main-results}

\begin{figure*}[!tbh]
    \centering
    \includegraphics[width=0.99\textwidth]{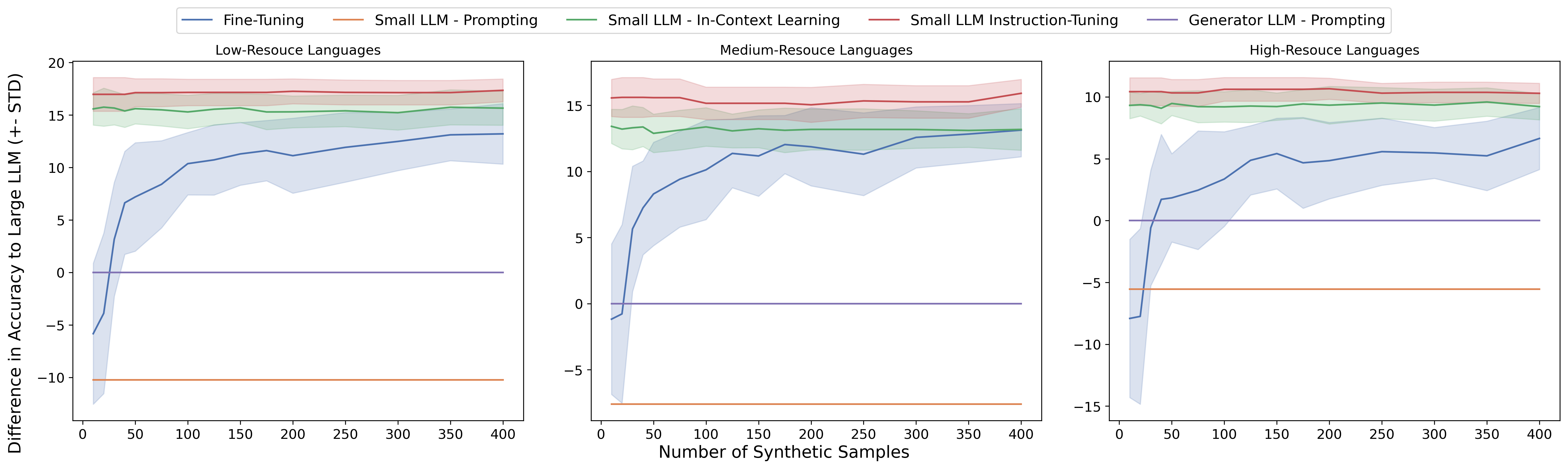}
    \caption{The difference in accuracy of smaller models trained using synthetic data compared to the large language model used for generating the data, aggregated across all tasks and language groups of different sizes. Using as few as $50$ synthetic samples, the smaller models achieve higher performance across all language groups.}
    \label{fig:main-comparison}
\end{figure*}

In this section, our goal is to answer the following main research question: \textit{\textbf{RQ1:} Can we distil the large language model by generating synthetic samples and using them for training smaller multilingual models to achieve more efficient classification performance on low-resource languages?} We also specifically focus on how many labelled samples we need to generate from the large language model in order to observe similar or better performance. The results for the LLaMA-3.1-8B model, aggregated across all tasks (excluding sarcasm detection), for low-resource, medium-resource, and high-resource languages are presented in Figure~\ref{fig:main-comparison}. The results for the remaining two smaller models (Gemma-3-4B, Qwen-2.5-7B) closely follow those observed for the LLaMA-3.1-8B model and, as such, are included in Appendix~\ref{app:language-comparison} (with a description of potential differences).

\textbf{Using generated synthetic samples, smaller models quickly outperform the large language model used for the generation.} To outperform the generator model, the smaller models require as few as $50$ synthetic samples. When taking the randomness (represented as standard deviation over the 20 runs) originating from different subsampling of the synthetic samples and training into consideration, we observe differences only for high-resource languages, where fine-tuning requires up to $100-150$ samples to always outperform the generator LLM. The overall increase in performance in comparison to this model, when using all the generated synthetic samples, is on average $10\%$ (percentual points). As can be expected, the highest performance is achieved when instruction-tuning the smaller language models. However, it is significantly more expensive than fine-tuning or inference through in-context learning, while not producing as much benefit.

\textbf{The benefit of distillation through synthetic samples depends on language groups}, with low-resource languages benefiting the most and high-resource languages the least. For low-resource languages, the increases are on average $13\%$ for fine-tuning, $15\%$ for in-context learning, and $18\%$ for instruction-tuning. For the high-resource languages, the increase drops to $5-6\%$ for fine-tuning, $9\%$ for in-context learning and $11\%$ for instruction-tuning. This difference can be expected, as high-resource languages (especially English) are overrepresented in all the large language models. 

Overall, we can conclude that the \textbf{large language models are better suited as generators of synthetic samples that are used to train smaller models than classifiers}, especially for low-resource languages. Such a setup not only leads to better performance but also more efficient classification, as we do not require as many computational resources. However, it is important to note that we \textbf{observe a significant impact of hyperparameter setup} (represented by the large standard deviation in the results in Figure~\ref{fig:main-comparison} due to failed runs, i.e., runs achieving performance worse than random chance) in the low-resource languages and when using synthetic samples -- especially due to the tendency of the fine-tuning models to quickly overfit on the synthetic samples. This effect can be especially observed when changing the number of samples we have available, such as in the low- or high-resource languages between $150$ and $200$ samples, where increasing the number of available samples can lead to a drop in performance.
\begin{figure*}[!tbh]
    \centering
    \includegraphics[width=0.99\textwidth]{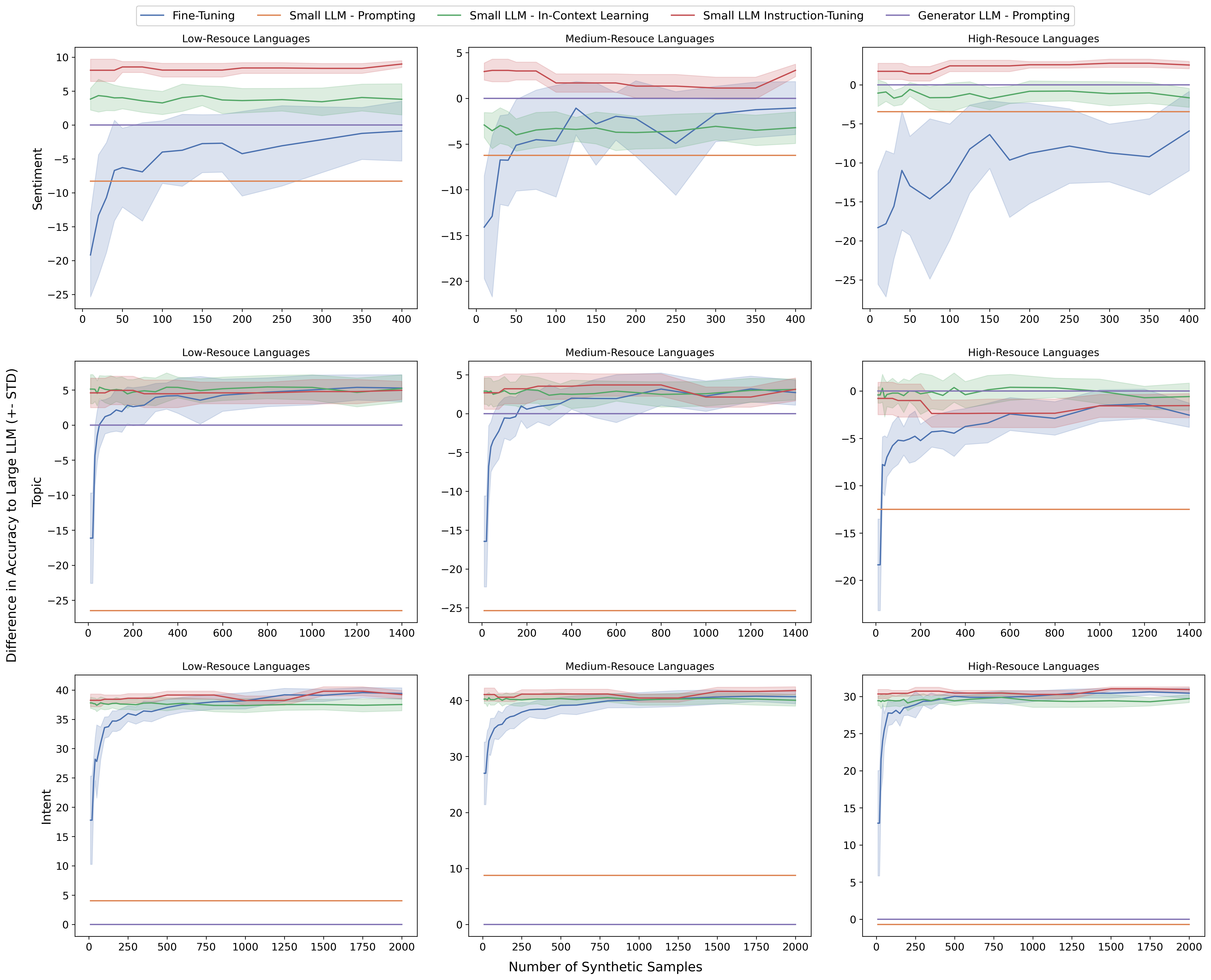}
    \caption{The comparison of the accuracy difference for tasks with different characteristics. For the commonly used sentiment classification task, we can observe significantly lower benefit of synthetic samples, especially for fine-tuning.}
    \label{fig:task-comparison}
\end{figure*}

\subsection{Impact of Task Characteristics}

In this section, we answer the following research question \textit{\textbf{RQ2:} How do the task characteristics affect the training using synthetic samples?} We analyse and compare the training across tasks with different characteristics, such as their difficulty, text lengths or how represented they are in the training data used in LLM pretraining. The results for the LLaMA-3.1-8B model are presented for the low-resource, medium-resource and high-resource languages in Figure~\ref{fig:task-comparison}. The results for the remaining models are included in Appendix~\ref{app:task-characteristics}.

\textbf{Task characteristics have a significant effect on the success of using synthetic training samples.} For the popular sentiment classification task (which is characterised by the lower difficulty due to binary classification and over-representation in pretraining data), we observe that fine-tuning on synthetic samples cannot outperform the large model used for generating the samples. For the low-resource languages, the performance only approaches the generator model when using all of the $400$ synthetic samples, while for the high-resource languages, we observe the performance of fine-tuning being lower by $5\%$ than the generator model. This follows the findings from previous works that found that fine-tuning on overly represented tasks (sentiment classification) requires a large number of samples to outperform classification with LLMs~\citep{pecher2024comparing}. Similarly, in the case of medium and high-resource languages, even using synthetic in-context examples with a smaller LLM does not lead to better performance. The only consistently better approach that utilises synthetic samples is instruction-tuning. However, it also incurs significant computation costs, which can often match those from using the large generator LLM.

On the other hand, for tasks characterised by a larger number of classes, longer sentences, and lower representation in pretraining data (topic and intent classification), we observe a large benefit of synthetic samples. For the topic classification, using synthetic samples for fine-tuning, instruction-tuning, or as in-context examples, is not able to outperform the generator LLM only in the case of high-resource languages. For medium and low-resource languages, we still observe up to $5\%$ improvement in the performance, which is significantly (i.e., by up to $30\%$) higher than prompting with smaller LLMs. For the intent classification, which represents the "hardest" task (in terms of number of classes and representation), all of the smaller models utilising synthetic samples outperform the large generator model with as low as $10$ samples. On low-resource languages, we observe an increase in performance of up to $40\%$, while on high-resource languages up to $30\%$.

\begin{figure*}[!tbh]
    \centering
    \includegraphics[width=0.99\textwidth]{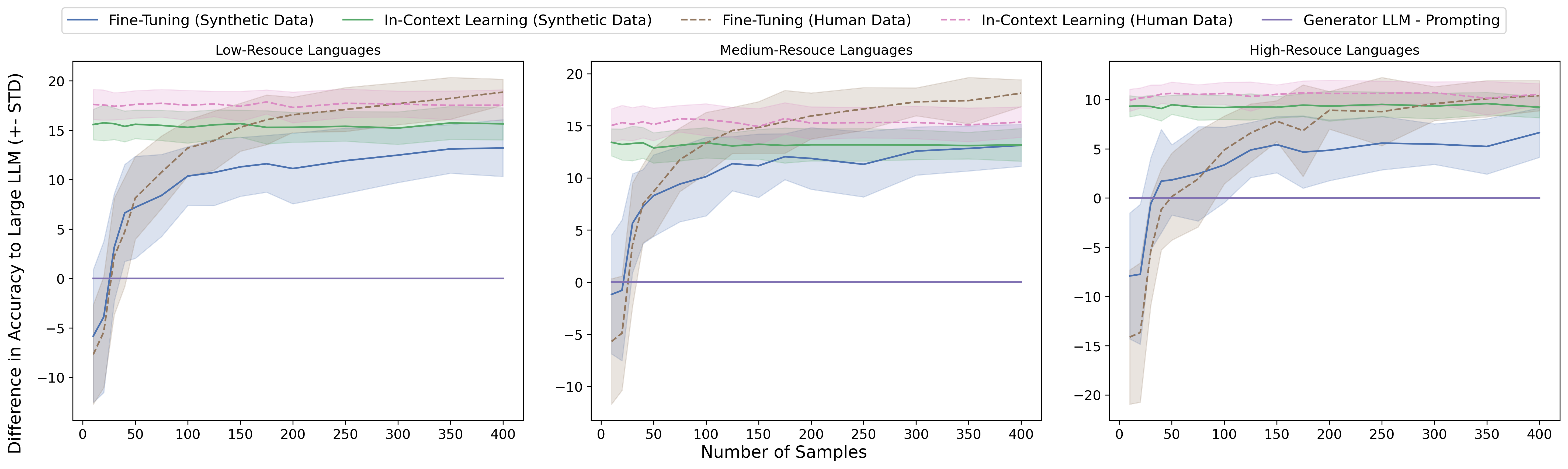}
    \caption{Comparison between synthetic and human labelled training samples. The human labelled samples provide more performance benefit, especially on larger number of samples.}
    \label{fig:synthetic-vs-human}
\end{figure*}

As such, we can conclude that \textbf{generating synthetic samples and using them as distillation is more effective for less represented tasks even in low-resource languages.} This indicates that the benefit does not depend only on the resources available for the languages, but the representation of tasks in the pretraining data or their popularity plays a strong role as well. To further confirm this finding, we run an additional analysis on a more niche task in Section~\ref{sec:niche-task}.

\subsection{Comparison Between Synthetic and Human Data}
\label{sec:synthetic-vs-human}

In this section, we focus on answering the following research question \textit{\textbf{RQ3:} How does the benefit of synthetic samples compare to human labelled data?} We use the human-labelled training samples from each language and task and use them to fine-tune the model or use them as in-context examples, following the same methodology (e.g., for subsampling, hyperparameters and overall training). We do not report results for the instruction-tuning as it incurs significant computation costs, while achieving similar performance to synthetic samples in our preliminary results. The results for the LLaMA-3.1-8B model aggregated over all tasks are presented for low-resource, medium-resource, and high-resource languages in Figure~\ref{fig:synthetic-vs-human}, and for the remaining models in Appendix~\ref{app:human-and-synthetic}.

\begin{figure*}[!tbh]
    \centering
    \includegraphics[width=0.99\textwidth]{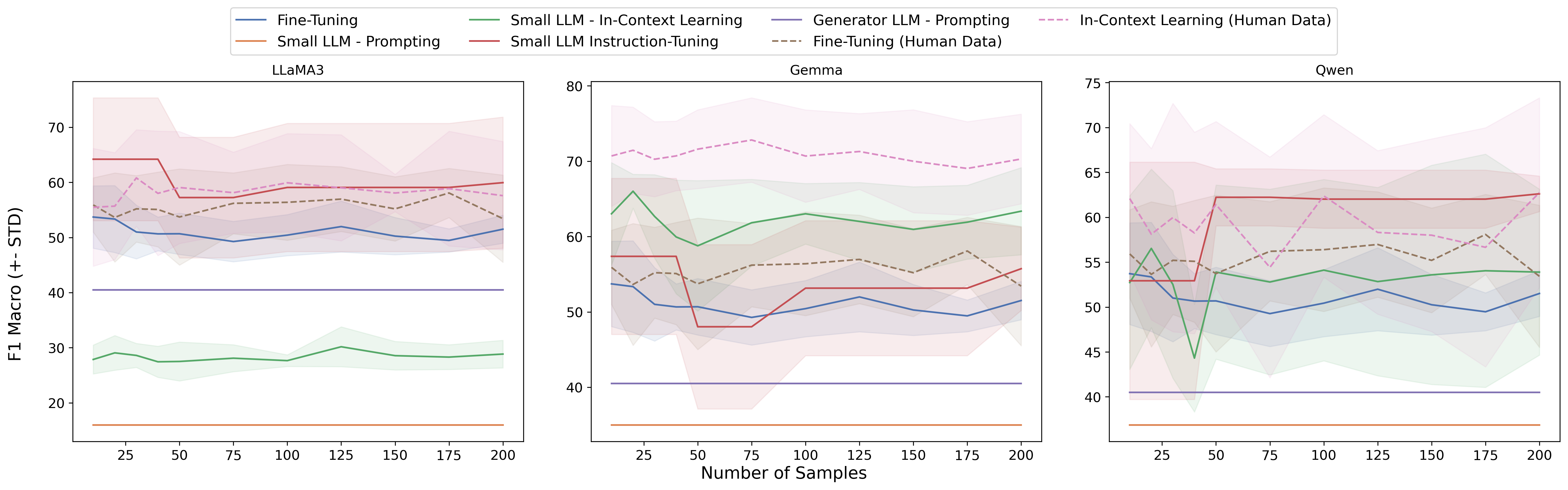}
    \caption{Using synthetic samples for sarcasm detection on English. On this more complicated task, the synthetic samples lead to performance improvements even on high-resource language.}
    \label{fig:sarcasm}
\end{figure*}

\textbf{Synthetic samples are comparable to human-labelled samples when it comes to the performance improvements only in resource-constrained settings.} When using only a small amount of samples for fine-tuning, up to $100$, we observe similar performance regardless of whether using synthetic or human-labelled data. However, as the number of samples increases, the performance from human-labelled data steadily increases, while for the synthetic samples, it starts to stagnate. For low- and medium-resource languages, the gap between synthetic and human data is up to $10\%$ when using all 400 samples, whereas for high-resource languages, the gap is only up to $5\%$. For in-context learning, we observe a consistent benefit of human data over synthetic data, with a smaller gap of around $4\%$ for low- and medium-resource languages and around $ 2\%$ for high-resource languages. This follows the findings from previous works that found that human-labelled samples provide larger diversity~\cite{guo2024benchmarking, reinhart2025llms} and may also explain the hyperparameter sensitivity (e.g., tendency to overfitting) when training on synthetic samples. However, it is important to note that the hyperparameter sensitivity and overfitting is observed also for the human-labelled samples (with the hyperparameters suited for synthetic samples being the best also for human-labelled ones). As the number of samples increases, the situation is improving slightly in both cases (human and synthetic samples). At the same, we do not employ any diversity incentives when generating samples, which may be the cause of lower performance benefits when using a larger number of synthetic samples (as compared to human-labelled ones) and using such incentives could help in these cases~\citep{cegin2024effectsdiversityincentivessample}. As such, this further reinforces the finding that distillation using synthetic samples (as is) should be used mostly in resource-constrained settings (low-resource languages and niche tasks) where getting human-labelled samples is problematic. For less resource-constrained settings, more attention to sample diversity should be given to potentially observe similar benefits.

Finally, \textbf{human-labelled samples lead to lower deviation in results than synthetic samples.} For the fine-tuning with human data, we observe a significantly lower number of failed runs and reduced sensitivity to hyperparameter setup and other sources of randomness.

\subsection{Additional Analysis: Low-Resource Task on High-Resource Language}
\label{sec:niche-task}

In this section, we extend our analysis of low-resource settings to sarcasm detection in English, a more complicated task characterised by lower availability of samples for training (and lower representation in pretraining samples). Our main focus is to investigate whether the benefit of synthetic samples transfers from the low-resource languages also to a combination of a high-resource language and a more niche task. The results for all three small models are presented in Figure~\ref{fig:sarcasm} using F1 macro due to the test set containing a significant sample imbalance (e.g., it is possible to achieve more than $80\%$ accuracy by predicting only a single class).

\textbf{For high-resource language, using synthetic samples leads to the largest benefit when interested in more complicated or niche task.} We observe that fine-tuning can always outperform the large generator model, with a difference in F1 of $12\%$, even though using more samples does not lead to any noticeable increase in the overall performance. This may be caused by the inability of the generator to produce samples with sufficient informativeness. However, we observe the same behaviour for the human labelled, where the performance remains more or less stable, but higher by up to $4\%$ over synthetic samples. For other approaches, we observe more varied results. In the case of the LLaMA-3.1-8B model, using synthetic samples in in-context learning does not lead to better performance than the generator model, although for Gemma-3-4B and Qwen-2.5-7B, it does (and leads to performance increases of $5-25\%$ F1). On the other hand, using human-labelled samples leads to consistent improvements of up to $30\%$. Similar behaviour can be observed for the instruction-tuning as well.

Furthermore, we \textbf{notice a significantly higher variance in results} across all of the approaches. It is especially noticeable for the instruction-tuning and in-context learning approaches, where the difference can be as high as $20\%$ F1, just based on how the samples are selected. As such it further reinforces the finding that \textbf{when using synthetic samples, more focus should be dedicated to the hyperparameter setup}, such as choosing learning rate, batch size or how the samples are selected.

\section{Conclusion}

In this work, we perform a systematic and comprehensive analysis to investigate whether the large language models are better employed as classifiers or as generators of synthetic samples for training smaller models in low-resource classification. We perform a kind of data-driven distillation of a large multilingual language model, which is used to generate a set of synthetic samples that are then used to fine-tune or instruction-tune a smaller model, or used as in-context examples. Based on our experiments spanning 11 languages and 4 tasks, we show that only a small number of generated synthetic samples is enough for the smaller models to achieve performance comparable to, and often surpassing, the large generator itself, especially in low-resource languages and less represented tasks. However, we also observe potential drawbacks with this approach, where the fine-tuning on the synthetic samples shows a significantly stronger sensitivity to the hyperparameter setup. At the same time, the synthetic samples lead to improvements comparable to human-labelled data only in settings with low number of samples (around $50$), with their benefit quickly stagnating due to their limited diversity and informativeness. As such, our findings suggest that the large language models are best utilised as generators in low-resource settings or for the purpose of quick prototyping, producing synthetic data that effectively distils their knowledge into smaller, more efficient models for practical use, rather than classifiers. This paradigm reduces the reliance on large models at inference time and offers a scalable path toward NLP in resource-constrained settings. As a possible future extension, we see a benefit of a more sophisticated generation strategy that would provide samples with higher diversity.

\section*{Acknowledgments}
This work was partially funded by the EU NextGenerationEU through the Recovery and Resilience Plan for Slovakia under the project \textit{AI-Auditology}, No. 09I03-03-V03-00020.

Part of the research results was obtained using the computational resources procured in the national project funded by the Ministry of Education, Youth and Sports of the Czech Republic through the e-INFRA CZ (ID:90254).

\section*{Limitations}

Due to the large scope of experiments in our study, focusing on multiple approaches and 11 different languages, which leads to us performing more than 30,000 fine-tuning and instruction-tuning runs, we limit the number of models used for generation to 1, fine-tuning models to 1, smaller LLMs to 3, and the number of tasks to 4. We try to mitigate this limitation by selecting the best-performing models through a preliminary study and focusing on well-established tasks.

Although we work in a setting with low-resource languages and use synthetic samples only, to generate the synthetic samples, we utilise a few human examples to drive the generation. Such an approach was found to produce significantly better examples, but may lead to samples similar to the datasets we are trying to replace. In addition, it also leads to potentially lower diversity of the samples, which we observe during our experiments. These issues could be solved by using approaches that drive the generation to create more diverse samples.

We use the same prompt for all the models, which is a result of prompt engineering on the LLaMA-3.1 model. The prompt was created based on the dataset description, the prompts used in related works, the formats recommended for different models (e.g., taking inspiration from~\cite{sun2023pushing}), and our own experience. As such, this may not represent the optimal format for all the models (as identified in previous works~\cite{sclar_quantifying_2023, weber-etal-2023-mind, pecher2023effects}), and performing the investigation on multiple different prompts may improve the overall model performance and affect the findings. However, we opted for using only a single optimised prompt format to reduce the computation costs.

\section*{Ethical Considerations}
The experiments in this paper work with publicly available datasets, citing the original authors. We do not work with any personally identifiable information or offensive content and perform no crowdsourcing for further data annotation. In addition, we are not aware of any potential ethical harms or negative societal impacts of our work, apart from the ones related to the advancement of the field of machine learning. We follow the license terms for all the models we use (such as the one required for the use of the LLaMA-3 models). It is possible that the large language models we use contain biases and potentially offensive or harmful content. However, the original authors of these models reduce this bias as much as possible. At the same time, although we generate synthetic samples and include them as part of the supplementary material, we believe that no problematic samples are contained due to the tasks we have chosen. Finally, as the highest impact of our study is the CO2 generated as part of our experiments, we publicly release the impact statement and the number of computers used below.

\paragraph{Impact Statement: CO2 Emissions Related to Experiments} The experiments presented in this paper used significant computational resources as they required multiple training and evaluation runs of multiple models (to deal with variance in results), as well as using large language models that require a lot of computation even just for the inference. Overall, the experiments were conducted using a private infrastructure, which has a carbon efficiency of 0.432 kgCO$_2$eq/kWh (default value used as the actual efficiency of our HW instance was not measured). A cumulative of 1000 hours of computation was performed on hardware of type A100 PCIe 80GB (TDP of 250W). Total emissions are estimated to be 108 kgCO$_2$eq of which 0 percents were directly offset. This includes all of the generation, training, and inference with all models, and the preliminary experiments. These estimations were conducted using the \href{https://mlco2.github.io/impact#compute}{MachineLearning Impact calculator} presented in \cite{lacoste2019quantifying}.

Whenever possible, we tried to reduce the compute resources used as much as possible. Most computational resources were used by the instruction-tuning of the smaller LLMs. As such, we decided to perform this operation with LoRA, while also limiting the number of steps, to reduce the cost as much as possible. Similarly, we do not perform the experiments across various large language models for generation, for fine-tuning, and as small LLMs, but instead choose only the best performing ones, to limit the number of training and evaluation runs we need in this paper, while also providing a good enough scope.

\bibliography{ref}

\appendix

\section{Experimental Setup: Further Details}
\label{app:further-details}
In this Appendix, we provide further details on our experimental setup.

\paragraph{Languages and Datasets}  We use 11 typologically diverse languages belonging to different groups based on their sizes: 1) low-resource, including Azerbaijani, Romanian, Slovenian, Telugu, and Welsh; 2) medium-resource, including Hebrew, Indonesian, Swahili, and Thai; and 3) high-resource, including English and German. For each language and task, we generate 200 samples per class, which results in 400 samples for sentiment classification, 1400 for topic classification, 2000 for intent classification, and 200 for sarcasm detection (where we generate only 100 samples per class). When subsampling the datasets, we start with overall 10 samples (ensuring we have at least 2 samples per class) and increasing the number of available samples until the full dataset -- the exact subsamples are 10, 20, 30, 40, 50, 75, 100, 125, 150, 175, 200, 250, 300, 350, 400, 500, 600, 800, 1000, 1250, 1500, 1750, 2000. As we perform the subsampling by randomly choosing the same number of samples from all classes, we repeat all of the experiments 20 times to deal with any effects of randomness.

\paragraph{Models and Training} For fine-tuning, we use the XLM-RoBERTa large model from huggingface\footnote{\url{https://huggingface.co/FacebookAI/xlm-roberta-large}}. For the small LLMs, we use LLaMA-3.1 8B\footnote{\url{https://huggingface.co/meta-llama/Llama-3.1-8B-Instruct}}, Gemma-3 4B\footnote{\url{https://huggingface.co/google/gemma-3-4b-it}} and Qwen-2.5 8B\footnote{\url{https://huggingface.co/unsloth/Qwen2.5-7B-Instruct}} all instruction tuned versions. For the large generator model, we run only zero-shot classification through prompting, while for the smaller LLMs, we also include in-context learning. The prompts used for all LLMs are included in Table~\ref{tab:prompt-format}. For in-context learning, we also evaluate different numbers of in-context examples per class, starting with 2 and increasing them up to 100, always reporting only the best performance. However, we observe that in almost all cases, the best performance is achieved using 2 samples per class (and for some, using 4). For instruction-tuning, we use the same prompt format and utilise LoRa~\citep{hu2022lora}, which is a method of parameter-efficient fine-tuning. We set the LoRA rank to 16, alpha to 16, and dropout to 0.05. We reparametrize each key, value, and downsample weight matrix of the transformer architecture. We train the model for 600 steps with a learning rate of 3e-5, with 10 warmup steps and batch size of 4 with early stopping. After training the model, we perform inference using in-context learning. For all of prompting and in-context learning inference, we set the LLMs to produce deterministic output (by setting temperature and turning off sampling) with up to 20 samples. For fine-tuning, we observe a significant sensitivity to the hyperparameters, especially across different dataset subsamples. As such, we run a hyperparameter optimisation for each language and dataset size and, to guarantee fairness in comparison, choose the setup that works the best across all of the experimental setups. We use a dropout of 0.3 in the classification layer, a learning rate of 1e-5, an AdamW optimiser with warmup, and train for 50 epochs with early stopping. We dynamically change the batch size across different dataset sizes, starting with a batch size of 4 at 10-20 samples, 8 at 20-50, 12 at 50-100, 16 at 100-150, 32 at 150-250, and 64 for all other sizes. Without this setup, we have observed a severe overfitting of the fine-tuning models.

\paragraph{Generation} For generating the synthetic samples, we use Llama-3-70b-instruct\footnote{\url{https://huggingface.co/TechxGenus/Meta-Llama-3-70B-Instruct-GPTQ}}. The generation prompt is included in Table~\ref{tab:prompt-format}. The \textit{task\_text\_type} placeholder had values based on tasks, either semantic analysis, intent recognition, topic classification, or sarcasm detection. The \textit{sep\_token} was represented as "---". The \textit{label} placeholder was replaced for each task and label with an LLM-generated explanation of that label based on randomly preselected human examples. The \textit{examples} placeholder contained the human examples for the given label being generated. We use the following parameters for generation \textit{temperature=0.7,  top\_p=0.9, max\_tokens=4096, repetition\_penalty=1.2}. We collected 10 generated samples per inference run for increased efficiency and performed cleaning of the generated data. We excluded duplicates and collected until 200 unique texts were generated.

\begin{table*}[!tbh]
\centering
\small
\caption{Prompt format for the large language model used for generating the synthetic samples and for the classification using prompting or in-context learning. The parts included in "\{\}" are replaced by the relevant data, such as task values or examples..}
\label{tab:prompt-format}
\begin{tabularx}{\textwidth}{@{}lX@{}}
\toprule
\textbf{Task Type} & \textbf{Prompt Format Text} \\ \midrule
 Generation   & Please create 6 different \{\textit{task\_text\_type}\} in the \{\textit{language}\} language, separated by the {\textit{sep\_token}} token. The \{\textit{task\_text\_type}\} should be about the \{\textit{label}\}. Note that some examples from the dataset look as follows: Examples: \{\textit{examples}\}. Output only the text in \{\textit{language}\} and nothing else! Do not number the texts! \\ \midrule \midrule 

Sentiment &  Determine the sentiment of the sentence using the following options: 1) negative; 2) positive. Use only these two options. \\
Topic &  Determine the topic of the sentence using the following options: 1) science and technology; 2) travel; 3) politics; 4) sports; 5) health; 6) entertainment; 7) geography. Use only these options. \\
Intent &  Determine topic of the sentence using the following options: 1) alarm query; 2) audio volume down; 3) calendar remove; 4) cooking recipe; 5) datetime convert; 6) send email; 7) play audiobook; 8) movie recommendation; 9) transport ticket; 10) weather query. Use only these options. \\
Sarcasm &  Determine whether the comment in the sentence is sarcastic using the following options: 1) serious; 2) sarcastic. Use only these two options. \\

\bottomrule

\end{tabularx}
\end{table*}

\section{Results for Remaining Models}

In this Appendix, we present the results of the experiments from the main part of the paper for the remaining models, specifically Gemma-3-4B and Qwen-2.5-7B, and draw a comparison with the results from the main content of the paper.

\subsection{Comparison Across Language Groups}
\label{app:language-comparison}

\begin{figure*}[tbh]
    \centering
    \includegraphics[width=0.99\textwidth]{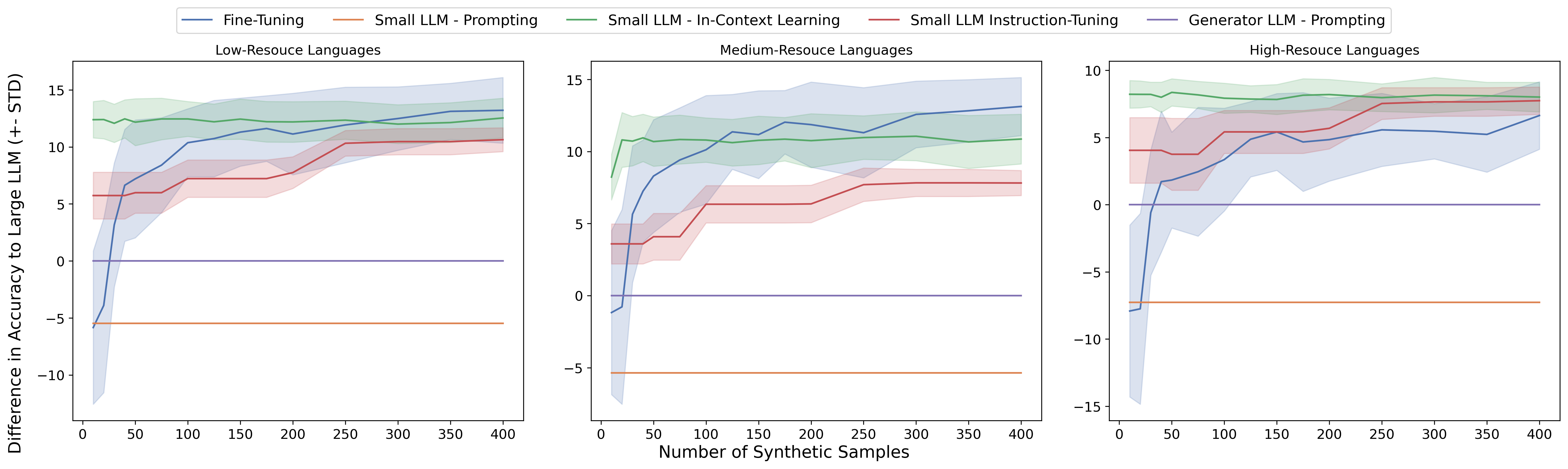}
    \caption{The difference in accuracy of smaller models (for Gemma model) trained using synthetic data compared to the large language model used for generating the data, aggregated across all tasks and language groups of different sizes.}
    \label{fig:gemma-comparison}
\end{figure*}

\begin{figure*}[tbh]
    \centering
    \includegraphics[width=0.99\textwidth]{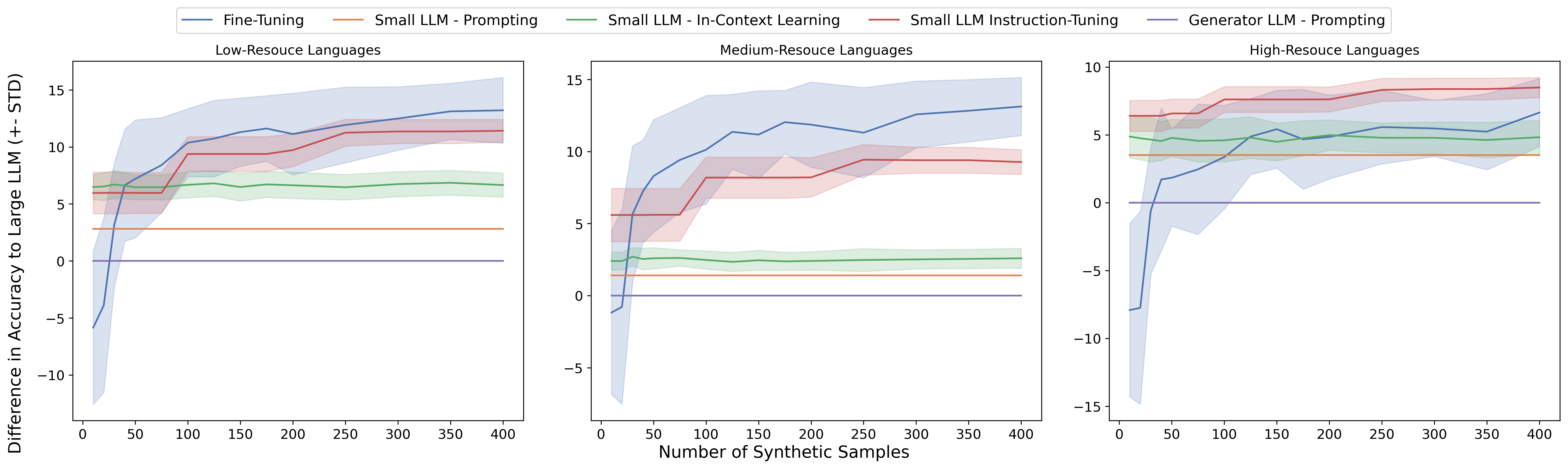}
    \caption{The difference in accuracy of smaller models (for Qwen model) trained using synthetic data compared to the large language model used for generating the data, aggregated across all tasks and language groups of different sizes.}
    \label{fig:qwen-comparison}
\end{figure*}

In this section, we present the results from the main comparison, aggregated across all tasks and languages, for the Gemma (Figure~\ref{fig:gemma-comparison}) and Qwen (Figure~\ref{fig:qwen-comparison}) models. Although the results follow the ones from the main part of the paper, there are some slight differences. The smaller LLMs still benefit from using synthetic samples, but we observe smaller performance improvements as compared to the LLaMA3 model. For the Gemma model, the improvement is up to $12\%$ for in-context learning and $10\%$ for instruction tuning. For the Qwen model, the improvement is up to $7\%$ for in-context learning and $10\%$ for instruction tuning. The difference may be explained by the lower overall performance capabilities of these models, although when looking at the results from zero-shot prompting, these models often perform better than LLaMA3. As such, these models may benefit less from the synthetic samples due to the discrepancy between the generator model and smaller LLMs -- the samples generated by the large LLaMA model may represent a different distribution for the Gemma and Qwen models, and thus provide lower benefit.

\subsection{Impact of Task Characteristics}
\label{app:task-characteristics}

\begin{figure*}[tbh]
    \centering
    \includegraphics[width=0.99\textwidth]{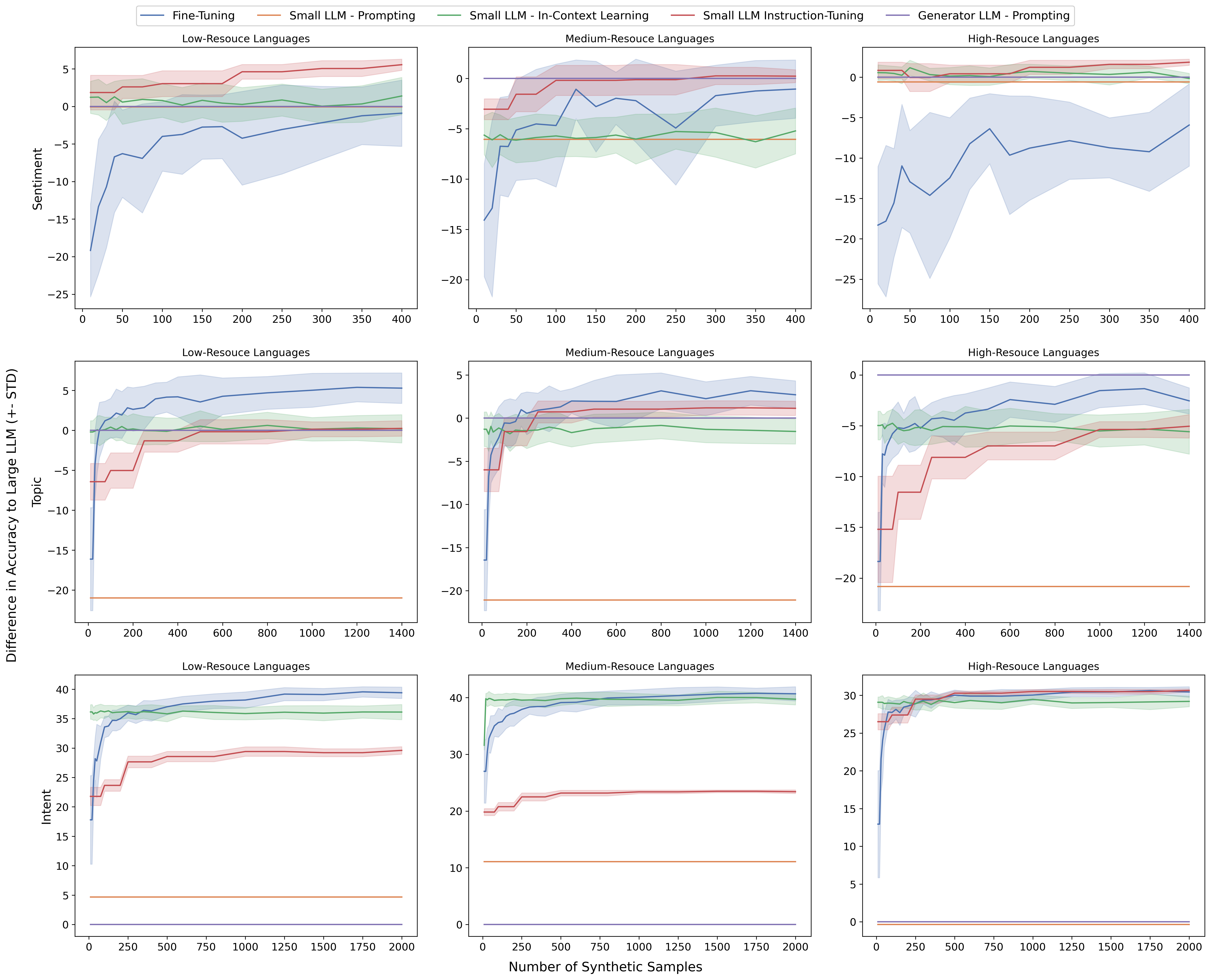}
    \caption{The comparison of the accuracy difference for tasks with different characteristics for the Gemma model.}
    \label{fig:gemma-task-comparison}
\end{figure*}

\begin{figure*}[tbh]
    \centering
    \includegraphics[width=0.99\textwidth]{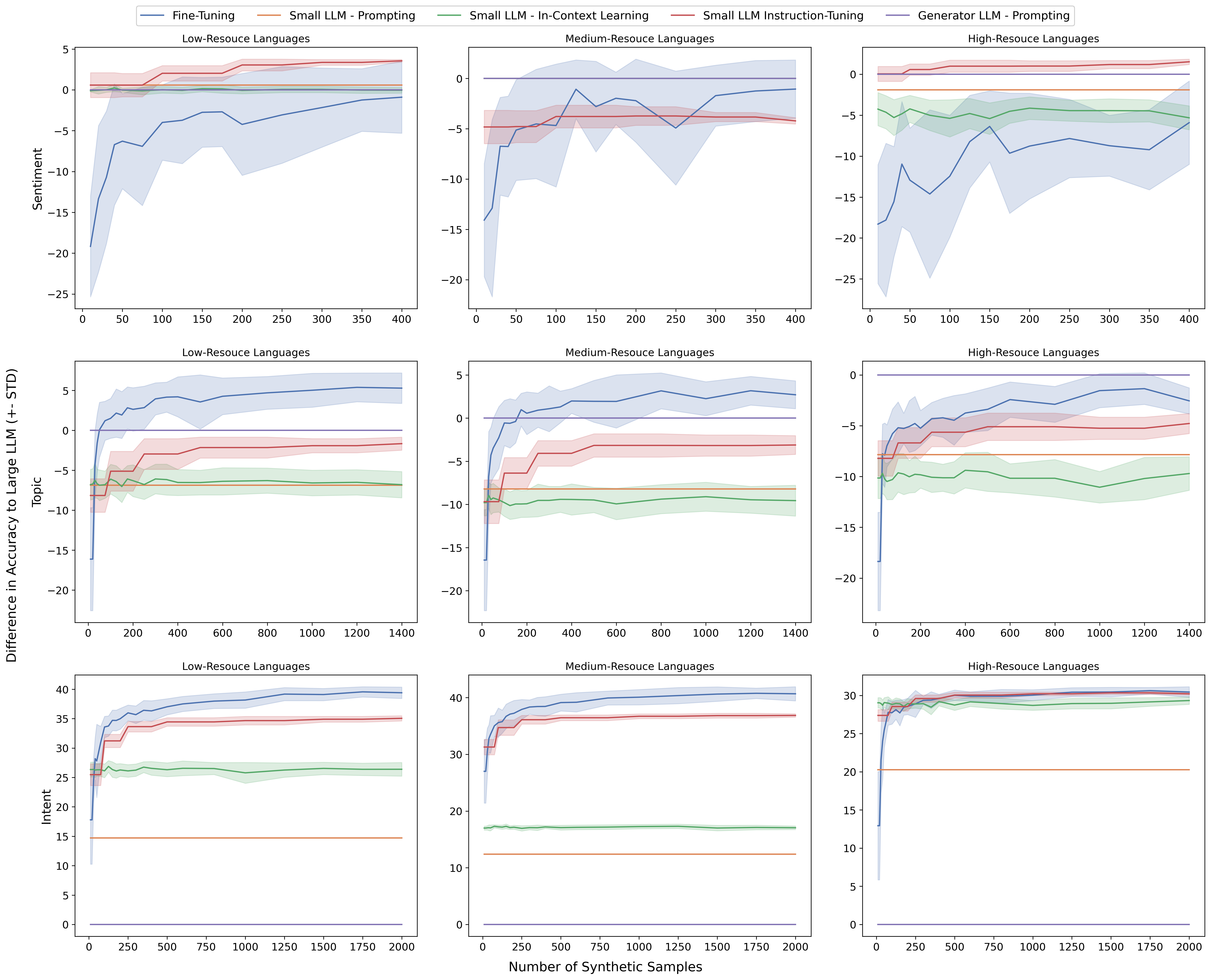}
    \caption{The comparison of the accuracy difference for tasks with different characteristics for the Qwen model.}
    \label{fig:qwen-task-comparison}
\end{figure*}

In this section, we present the results for the impact of task characteristics, aggregated for the different language groups, for the Gemma (Figure~\ref{fig:gemma-task-comparison}) and Qwen (Figure~\ref{fig:qwen-task-comparison}) models. In comparison to the results from the main content of the paper, we observe a lower benefit of the synthetic samples for the smaller LLMs used through in-context learning or instruction-tuning. The smaller models fail to outperform the generator model only in 4 cases out of 18, with both models on the sentiment task for medium-resource languages and the topic task or high-resource languages. For intent task, we do not observe such behaviour. At the same time, they often underperform the fine-tuning with synthetic samples. This follows the findings from the previous section and, as such, may be explained by the worse overall performance of these models (caused by lower number of parameters) or the discrepancy between the generator and these classification models (the samples generated by the LLaMA generator model may be better suited for the smaller LLaMA architectures as compared to different model families). However, it is important to note that the Gemma model benefits more from an increased number of samples when it comes to instruction tuning, which we do not observe for the other models. As such, it may indicate that this model has a larger room for improvement, as it is also the smallest LLM we use. Especially for the high-resource languages, the model can be improved by up to $10-15\%$ through instruction tuning on the synthetic samples.

\subsection{Comparison Between Synthetic and Human Data}
\label{app:human-and-synthetic}

\begin{figure*}[tbh]
    \centering
    \includegraphics[width=0.99\textwidth]{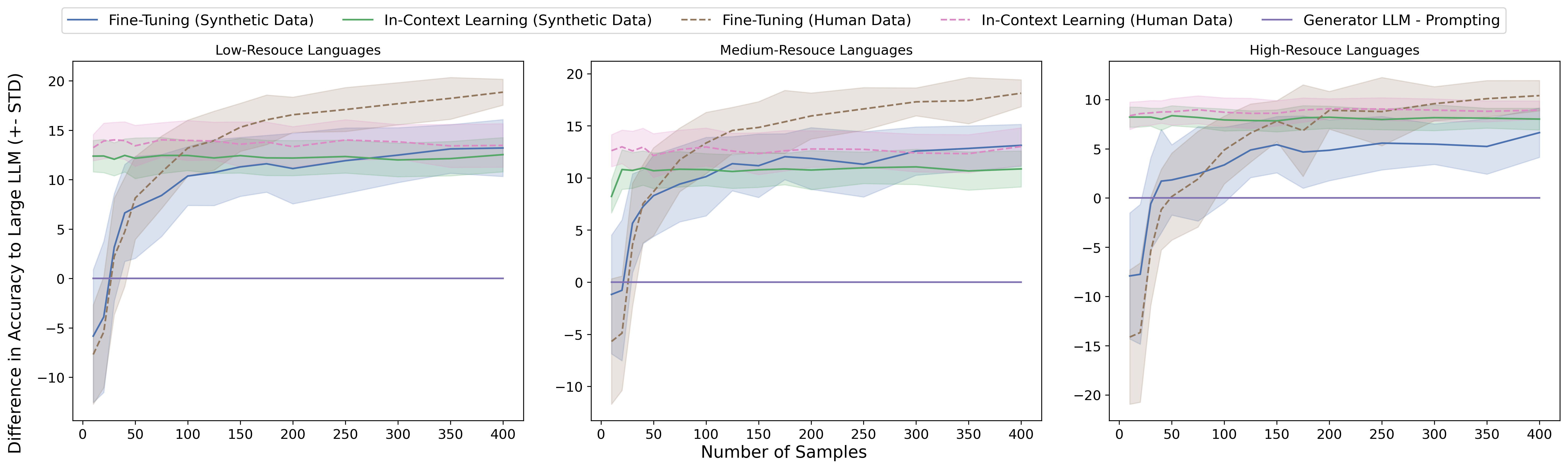}
    \caption{Comparison between synthetic and human labelled training samples for the Gemma model.}
    \label{fig:gemma-synthetic-vs-human}
\end{figure*}

\begin{figure*}[tbh]
    \centering
    \includegraphics[width=0.99\textwidth]{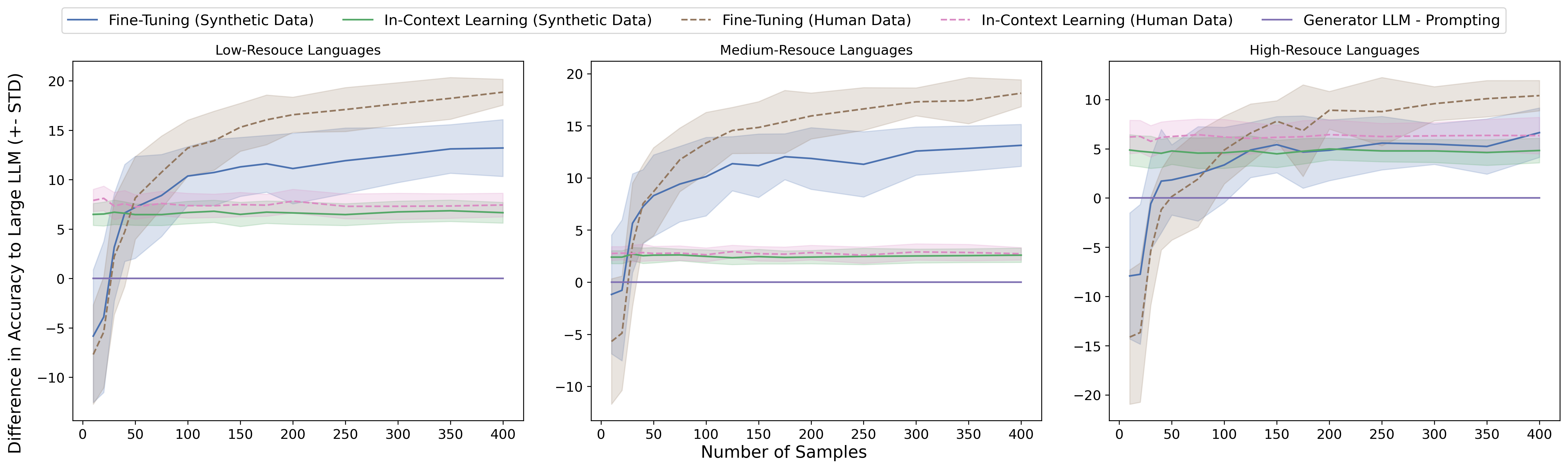}
    \caption{Comparison between synthetic and human labelled training samples for the Qwen model.}
    \label{fig:qwen-synthetic-vs-human}
\end{figure*}

In this section, we present the results from comparing synthetic and human data for the Gemma (Figure~\ref{fig:gemma-synthetic-vs-human}) and Qwen (Figure~\ref{fig:qwen-synthetic-vs-human}) models. As opposed to the results from the main content of the paper, we observe a significantly lower difference between synthetic and human data, with the difference as high as $2\%$ for both the Gemma and Qwen models. This observation may also provide an explanation for the results in the previous sections, as the models benefit less from additional samples and may be less suited for in-context learning and further instruction tuning. As with the previous sections, the models are often still outperformed by simple fine-tuning.

\section{Computational Costs}

In this appendix, we provide a summary of the running time (determined as GPU hours) for a single run, including inference and training if necessary, on the same GPU for the models. The summary is provided in Table~\ref{tab:gpu-costs}. As the number of samples that is considered for inference increases, the cost for the generator model increases at a significantly higher rate compared to other models. At the same time, the cost for instruction-tuning is mostly for the training. We can see that using smaller models is more efficient in the long run when it comes to GPU hours required.

\begin{table}[tbh]
\centering
\small
\caption{Comparison of costs, calculated as the average GPU wall-clock time for a single full run, including inference and training if necessary. }
\label{tab:gpu-costs}
\begin{tabular}{lc}
\textsc{Model}             & \textsc{GPU time (h)}  \\ \toprule
Generator Model & 15      \\
XLM-RoBERTa      & 0.167 \\
Prompting             & 1-2 \\
In-Context Learning                 & 3-4      \\
Instruction-Tuning        & 10      \\

\bottomrule
\end{tabular}

\end{table}

\end{document}